\documentclass[12pt]{article}
\pdfoutput=1
\usepackage{setspace}
\usepackage[top=1in, bottom=1in, left=1in, right=1in]{geometry}
\usepackage{enumitem} 
\setlist{nolistsep} 

\usepackage{amsmath,amssymb,amsthm,amsthm,amsfonts,latexsym,bbm}
\usepackage{amscd,amsxtra} 
\usepackage{graphics,graphicx,xcolor}
\usepackage{ifpdf}
\usepackage[caption=false]{subfig}
\usepackage{multirow}
\usepackage[colorlinks,linkcolor=blue,citecolor=blue,urlcolor=blue]{hyperref}
\usepackage[comma]{natbib}
\usepackage{comment}
\usepackage{ulem}
\usepackage{authblk} 
\usepackage{pifont} 
\newcommand{\cmark}{\ding{51}}%
\newcommand{\xmark}{\ding{55}}%

\allowdisplaybreaks

\def\ev{\boldsymbol e}
\def\fv{\boldsymbol f}
\def\gv{\boldsymbol g}

\def\uv{\boldsymbol u}

\def\xv{\boldsymbol x}
\def\yv{\boldsymbol y}

\def\Av{\boldsymbol A}

\def\Kv{\boldsymbol K}

\def\Rv{\boldsymbol R}

\def\Xv{\boldsymbol X}
\def\Yv{\boldsymbol Y}
\def\Zv{\boldsymbol Z}

\newcommand{\Cc}{\mathcal{C}}

\newcommand{\Lc}{\mathcal{L}}

\newcommand{\Sc}{\mathcal{S}}

\newcommand{\Yc}{\mathcal{Y}}

\newcommand{\alphav}{\mbox{\boldmath{$\alpha$}}}

\newcommand{\gammav}{\mbox{\boldmath{$\gamma$}}}

\newcommand{\zetav}{\mbox{\boldmath$\zeta$}}

\newcommand{\xiv}{\mbox{\boldmath{$\xi$}}}

\newcommand{\varphiv}{\mbox{\boldmath{$\varphi$}}}

\newcommand{\omegav}{\mbox{\boldmath{$\omega$}}}

\newcommand{\thetav}{\mbox{\boldmath{$\theta$}}}

\def\1v{\mathbf 1}
\def\0v{\mathbf 0}
\def\Id{\mathbf I} 

\newcommand{\Ind}[1]{\mathbbm{1}_{\left\{ {#1} \right\} }}

\newcommand{\R}{\mathbb R}

\newcommand{\E}{\mathbb E}
\newcommand{\sgn}{\mathop{\mathrm{sign}}}
\def\Pr{\mathbb P}

\newcommand{\diag}{\mathop{\rm diag}}

\newcommand{\mb}{\mbox}

\newcommand{\argmax}{\operatornamewithlimits{argmax}}

\newcommand{\set}[1]{\left\{#1\right\}}

\def\ie{\textit{i.e.}}

\theoremstyle{plain}
\newtheorem{theorem}{\sc Theorem}

\theoremstyle{remark}

\theoremstyle{definition}
\newtheorem{definition}{\sc Definition}

\title{Noncrossing Ordinal Classification}
\author{Xingye Qiao}
\affil{Department of Mathematical Sciences\authorcr  State University of New York, Binghamton, NY 13902-6000.\authorcr E-mail: \texttt{qiao@math.binghamton.edu}}
\date{}

\begin{document}
\setlength{\belowdisplayskip}{2pt} \setlength{\belowdisplayshortskip}{2pt}
\setlength{\abovedisplayskip}{2pt} \setlength{\abovedisplayshortskip}{2pt}
\maketitle
\pagenumbering{Roman}

\begin{abstract}
Ordinal data are often seen in real applications. Regular multicategory classification methods are not designed for this data type and a more proper treatment is needed. We consider a framework of ordinal classification which pools the results from binary classifiers together. An inherent difficulty of this framework is that the class prediction can be ambiguous due to boundary crossing. To fix this issue, we propose a noncrossing ordinal classification method which materializes the framework by imposing noncrossing constraints. An asymptotic study of the proposed method is conducted. We show by simulated and data examples that the proposed method can improve the classification performance for ordinal data without the ambiguity caused by boundary crossings.

\vspace{0.5in}

\end{abstract}

\newpage
\pagenumbering{arabic}
\setcounter{page}{1}

\section{Introduction}\label{sec:intro}

Data with ordinal class labels are very common in reality and they are collected from many scientific areas and social practices, such as disease diagnosis and prognosis, national security threat detection, and quality control. For example, the development of tumor can be classified to Stage I, Stage II, Stage III, \textit{etc.}; the U.S. homeland security advisory system has five categories, Green, Blue, Yellow, Orange and Red, ordered from the least to the most severe threats; the quality of a randomly sampled product can be categorized to excellent, good, fair and bad. The goal of ordinal classification is to classify a data point to one of these ordinal categories, $y\in\Yc$, based on the covariates $\xv\in\Sc\subset\R^d$. Here we consider the case $\Yc=\set{1,2,\dots,K}$. The actual labels are of no importance, so long as the order can be recognized.

Note that ordinal data are a special case of the more general multicategory data. Ignoring the order information, one may classify ordinal data in the same way as one would do multicategory data, by applying a multicategory classification method. There is a large body of literature for the latter. This includes those which combine multiple binary classifiers, such as the One-Versus-One and One-Versus-Rest paradigms \citep[see for example][]{duda2001pattern}, and those which estimate multiple classification boundaries simultaneously, such as \citet{Weston99support}, \citet{Crammer2002on}, \citet{lee2004multicategory}, and \citet{Huang2013Multiclass}. While using multicategory classification method for ordinal data sometimes works, such treatment can be suboptimal, because the classes are treated equally without their connections and relative superiority being considered. Moreover, a counterexample in Section \ref{sec:motivation} reveals that it is desirable to use an approach which fully utilizes the ordinal information available.


Suppose there are $K$ classes in total. A simple but very useful strategy for ordinal classification is to sequentially conduct binary classifications between the combined meta-class $\Cc_k\equiv \set{1,\dots,k}$ and meta-class $\overline\Cc_k\equiv \set{k+1,\dots,K}$, for $1\leq k\leq K-1$, and then pool the classification results from these $(K-1)$ steps to reach a final prediction \citep[see][]{Frank2001Simple}. In binary classification, usually a discriminant function $f$ is estimated, and data point $\xv$ is classified to the positive class if $f(\xv)>0$, or to the negative class otherwise. The classification boundary is defined by $\set{\xv:f(\xv)=0}$. As there are $(K-1)$ binary classifiers in this strategy, there are $(K-1)$ classification boundaries. This approach assumes that each class is \textit{sandwiched} by two adjacent classification boundaries. 

An inherent difficulty of this approach is that since these boundaries are trained separately, it is possible that they may cross with each other. Consequently, how to make a final conclusion becomes ambiguous for some data points.

In this article, we propose a flexible margin-based classification method for ordinal data. The direction we pursue is to construct the $(K-1)$ boundaries simultaneously. Our method is equipped with extra noncrossing constraints to fix the crossing issue, hence is named \textbf{N}oncrossing \textbf{Ordi}nal \textbf{C}lassification (NORDIC). Similar noncrossing constraints were studied and used in the quantile regression context \citep[for example,][]{bondell2010noncrossing,liu2011simultaneous}. Compared to the vanilla idea of training $(K-1)$ binary classifiers separately, simultaneous learning can borrow the strength from different classes, which leads to better classification accuracy and improved robustness to mislabeled data. Moreover, compared to many existing methods, our method is more flexible, since it does not assume that the boundaries are parallel.

Among the existing related work in classifying ordinal data, \citet{herbrich1999large} tried to find the classification boundaries by maximizing the margin in the space of pairs of data vectors; \citet{Frank2001Simple} was among the first to consider the idea of pooling binary classifiers; \citet{shashua2003ranking} generalized the support vector formulation for ordinal regression and proposed to optimize multiple thresholds to define parallel separating hyperplanes; \citet{chu2005new} improved the work of \citet{shashua2003ranking} and guaranteed that the thresholds were properly ordered; \citet{chu2006gaussian} used a probabilistic kernel approach based on Gaussian processes; \citet{cardoso2007learning} replicated the data and cast the ordinal classification problem to a single binary classification problem. Many of these approaches, although ensuring noncrossing, have posed a fairly strong assumption that the $(K-1)$ boundaries are parallel  to each other (either in the original sample space or in the kernel feature space), which may be lack of flexibility and be unrealistic in many cases.

The rest of the article is organized as follows. In Section \ref{sec:motivation}, we compare the multicategory classification with the ordinal one, and review a simple framework for the ordinal classification. We introduce the main idea of the NORDIC method and the computation algorithm in Section \ref{sec:methodology}. A more precise version of NORDIC, which makes use of a less popular optimization algorithm, is introduced in Section \ref{sec:NORDICIP}. The theoretical properties are studied in Section \ref{sec:theory}.  Several simulated examples are used to compare NORDIC with other methods in Section \ref{sec:simulation}. A real data example is studied in Section \ref{sec:real}. Concluding remarks are made in Section \ref{sec:conclude}.

\section{Ordinal Classification}\label{sec:motivation}
In this section, we first demonstrate, using a real example that, in some cases, it is better not to ignore the ordinal information by treating ordinal data as regular multicategory data. We then introduce a framework of ordinal classification via binary classifiers. Lastly we compare the principles of multicategory and ordinal classifications.
\subsection{An Example in U.S. Presidential Election}
In a multicategory classifier with $K$ classes, usually $K$ discriminant functions $g_k(\xv)$, $k=1,\dots,K$, are estimated and the class prediction for $\xv$ is $\argmax_{k\in\set{1,\dots,K}}g_k(\xv)$.  Let $\eta_k(\xv)$ denote the conditional probability for the $k$th class, $\eta_k(\xv)=\Pr(Y=k|\Xv=\xv)$. In this case, any multicategory classifier would aim to mimic the Bayes classification rule, $\phi^{Bayes}_{MC}(\xv)\equiv \argmax_{k\in\set{1,\dots,K}}\eta_k(\xv)$, which has the smallest conditional classification risk, $\Pr(\phi(\Xv)\neq Y|\Xv=\xv)$, among all possible rules.

For the ordinal data, one can opt to ignore the ordinal information and classify them using a multicategory classifier. However, a counterexample suggests that this may not always be a wise strategy. Consider the presidential election in the United State. Any voter can be viewed as being from a red state (a state which is most conservative and predominantly vote for the Republican Party), a blue state (a state which is least conservative and predominantly vote for the Democratic Party) and a purple state (also known as a swing state, where both parties receive strong support). In 2012, the states of North Carolina, Florida, Ohio, and Virginia were the swing states. There are many more blue and red states in the U.S. than swing states (and a much larger population in the former two types of states than that in the latter). Suppose each voter is associated with a covariate vector $\xv\in\Sc$ and the color of her home state is the class label. The statistical task here is to classify her to one of the three types of states, $\Yc=\set{\mb{red,~purple,~blue}}$.

Recall that the Bayes rule in multicategory classification classifies $\xv$ to the class with the greatest $\eta_k(\xv)$. It is more likely for a multicategory classifier to classify a voter to a blue state or a red state, since both tend to have larger $\eta_k(\xv)$. To see this, note that $\eta_k(\xv) = \pi_kd_k(\xv)/\{\sum_{\ell\in\Yc}\pi_\ell d_\ell(\xv)\}$, where $d_k(\xv)$ is the density of the covariate $\Xv$ given that she is from the $k$th class and $\pi_k$ is the unconditional class probability for the $k$th class. Clearly, both $\pi_{\mb{red}}$ and $\pi_{\mb{blue}}$ are \textit{much} greater than $\pi_{\mb{purple}}$, leading to that their $\eta_k(\xv)$'s tend to be larger as well. The bottom line is, it seems to be unfair that the chance that a voter from the purple state is correctly identified is compromised simply because there is a smaller population in purple states. Ironically, in a U.S. presidential election, the swing states are the most important battleground, because it is the swing states that break the even in a presidential campaign. 

In this example, the imbalanced class prior probabilities appear to be the proximate cause that leads to the aforementioned issue. The underlying root cause, however, is that the ordinal data nature herein has been ignored. A classification method which makes use of the ordinal information is more appropriate in this case. We describe a simple strategy for this example here which leads to the more formal methodology in the next subsection: for a randomly selected voter, we first consider classifying her to a \textit{blue} state, versus a \textit{purple or red} state. If she is classified to the latter, then she tends to be relatively more conservative (than blue states voters). We then classify her to a \textit{blue or purple} state, against a \textit{red} state. If she is classified to the former, then she is relatively less conservative (than red state voters). The results of the two comparisons can lead to the final conclusion that she is classified to a purple state.


\subsection{Ordinal Classification via Binary Classifiers}
In general, consider an ordinal classification problem with $K$ classes. Furthermore, consider $(K-1)$ binary classifiers, where the $k$th classification boundary separates the combined set $\set{i:y_i\in\Cc_k}$ from the combined set $\set{i:y_i\in\overline\Cc_k}$ where $\Cc_k\equiv \set{1,\dots,k}$ and $\overline\Cc_k\equiv \set{k+1,\dots,K-1}$. For the $k$th binary classification, we code the former the negative class and the latter the positive class by constructing a dummy class label $y^{(k)}\equiv -1$ if $y\leq k$ and $+1$ if $y>k$. The $k$th binary classifier is associated with a discriminant function $f_k(\xv)$ so that the classification rule is $\sgn\{f_k(\xv)\}$. Let $Z_k(\xv)$ denote the prediction set of observation $\xv$ with respect to the $k$th subproblem, defined as $\Cc_k$ if $f_k(\xv)<0$, or $\overline\Cc_k$ otherwise. The final prediction for $\xv$, aggregating all the results from the $(K-1)$ binary classifiers above, will be the intersection of $Z_k(\xv)$, \ie, $\bigcap_{1\leq k\leq K-1} Z_k(\xv)$. 

\begin{table}[!htb]
	\centering
		\begin{tabular}{c|c|c|c}
			& BC I & BC II & BC III \\\hline
			Class 1 & \xmark & \cmark & \cmark \\\cline{2-2}
			Class 2 & \cmark & \cmark & \cmark \\\cline{3-3}
			Class 3 & \cmark & \xmark & \cmark \\\cline{4-4}
			Class 4 & \cmark & \xmark & \xmark
		\end{tabular}
			\caption{\small An illustrative table showing the predictions of the three binary classifiers for an observation in a four-class example. Aggregating the results of the three binary classifiers, we can reach the final prediction that the observation is classified to the second class. BC is short for ``Binary Classifier''.}\label{tab:example}
\end{table}

In a four-class toy example, Table \ref{tab:example} tabulates the prediction of the three binary classifiers for some  observation $\xv$. The first binary classifier compares Class 1 and the meta-class $\set{2,3,4}$. The prediction is that the observation is from $\set{2,3,4}$. Similarly, the second binary classifier compares $\set{1,2}$ and $\set{3,4}$ and the prediction is $\{1,2\}$. Lastly, the third binary classifier classifies the observation $\xv$ to $\set{1,2,3}$. Clearly, Class 2 is favored by all three binary classifiers and it is the final prediction for $\xv$. This framework for reaching an ordinal classification prediction by pooling binary classifiers was first noted by \citet{Frank2001Simple}.

\subsection{Principle of Ordinal Classification}
\begin{figure}[!b]
	\centering
		\includegraphics[width = 0.7\linewidth]{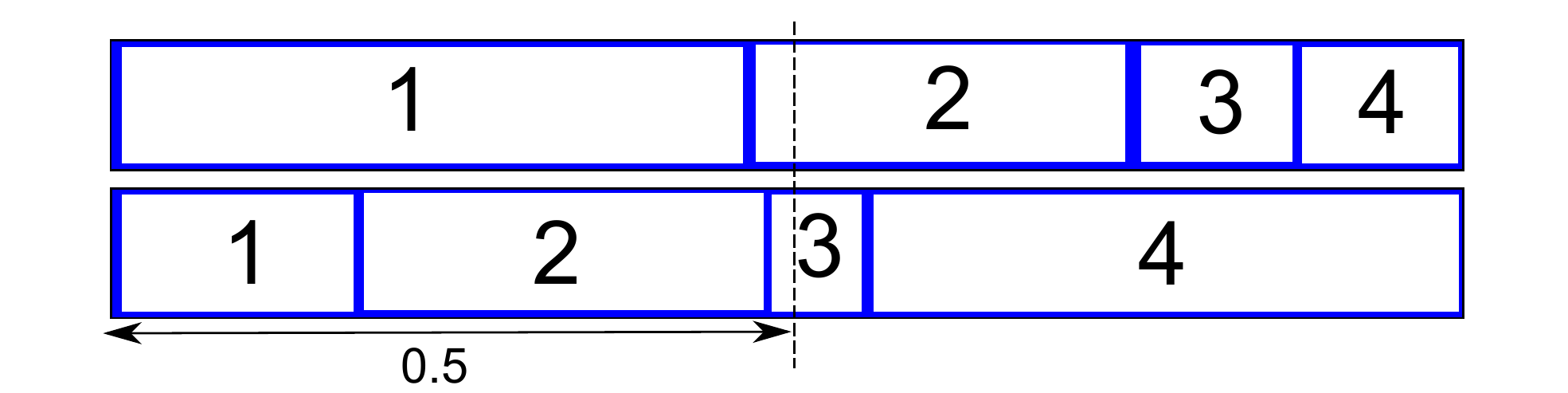}\vspace{-1em}
		\caption{\small In the top panel, the multicategory classification principle chooses Class 1 while the ordinal classification principle chooses Class 2. In the bottom panel, the multicategory classification principle chooses Class 4 while the ordinal classification principle chooses Class 3. }
	\label{fig:eta}
\end{figure}
We are now ready to compare the principles of multicategory classification and ordinal classification. A cartoon in Figure \ref{fig:eta} can tellingly demonstrate the distinction between these principles. In a data set with $K=4$, there are two example data points (shown in the top and the bottom rows respectively). For each data point, the length of each block denotes the conditional class probability $\eta_k(\xv)$. The sum of all four conditional probabilities is 1. The principle in multicategory classification chooses Class 1 in the top example and Class 4 in the bottom example, as they correspond to the greatest $\eta_k(\xv)$ in both cases. In contrast, in ordinal classification, the desired prediction would be Class 2 and Class 3 respectively. For example, for the top example, the data point is more likely from Class $\{1,2\}$ than from Class $\{3,4\}$, and more likely from  Class $\{2,3,4\}$ than from Class $\{1\}$. Hence Class 2 is the most plausible choice for this data point. Similarly, the data point in the bottom is most likely from Class 3. In particular, they both correspond to Class $k$ such that $\sum_{\ell=1}^{k-1} \eta_{\ell}(\xv)< 1/2$ and $\sum_{\ell=1}^{k}\eta_{\ell}(\xv)>1/2$ for each $\xv$. In the cartoon, a vertical line corresponding to 0.5 cuts the blocks for the desired predictions.

A useful notion here is that the principle of multicategory classification is to select the `mode' of the class labels, based on $\eta_k(\xv)$, while that of the ordinal classification is to select the `median'.

\section{Noncrossing Ordinal Classification}\label{sec:methodology}
Conducting ordinal classification via binary classifiers is very easy to implement as long as one has access to an efficient binary classifier. There are many options, such as Support Vector Machine \citep[SVM;][]{Cortes1995Support, vapnik1998statistical, cristianini2000introduction}, Distance Weighted Discrimination \citep[DWD;][]{marron2007distance,qiao2010weighted}, hybrids of the two \citep{Qiao2015Flexible,Qiao2015Distance}, $\psi$-learner \citep{Shen2003}, Large-Margin Unified Machines \citep{Liu2011Hard} and so on.

\begin{figure}[!htb]
	\centering
		\includegraphics[width=0.6\textwidth]{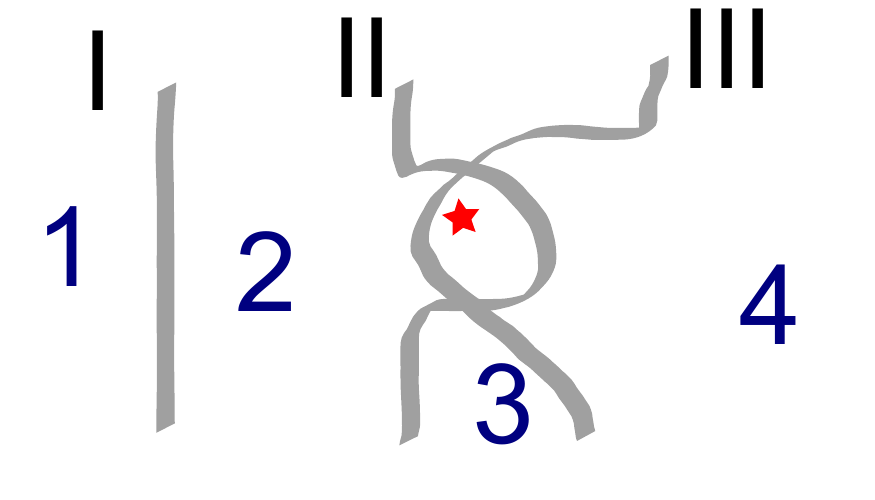}
		\caption{\small A cartoon showing the possible crossing between estimated classification boundaries. Four classes of data (annotated as 1,~2,~3 and 4) with three estimated classification boundaries (I, II and III). Their true noncrossing boundaries are implied by their locations and are not shown. The second and the third estimated boundaries cross with each other. Consequently, the red star point cannot be classified properly.}
	\label{fig:drawing_fig1}
\end{figure}

However, because the $(K-1)$ classification boundaries are trained separately, it is possible that they cross with each other. Figure \ref{fig:drawing_fig1} is a cartoon which shows the possible crossing between classification boundaries. Here there are four classes (annotated as 1,~2,~3 and 4) and three estimated classification boundaries (I, II and III). The second and the third estimated boundaries cross with each other. Consequently, the red star point cannot be classified properly. In particular, it will be classified by classifier I to $\set{2,3,4}$, by classifier II to $\set{1,2}$ and by classifier I to $\set{4}$. The intersection of all three prediction sets is empty. Although one may argue that this point might be Class 2 or Class 4, no definite answer can be given, and there is an ambiguity as to how to classify this red star point.

Hence, it is desired that the estimated classification boundaries do not cross with each other. Let $f_k(\xv)$ be the discriminant function for the $k$th binary classification. Recall that its boundary are defined by $\set{\xv:f_k(\xv)=0}$. For these boundaries to be noncrossing, mathematically, it is equivalent that for all $\xv\in\Sc$ not on any boundary, where $\Sc$ is a subset of $\R^d$, there exists $k\in\set{1,2,\dots,K-1}$, such that $f_{\ell}(\xv)>0$ for all $\ell<k$ and $f_{\ell}(\xv)<0$ for all $\ell\geq k$. Let $S(\xv,k)\equiv \sgn\{f_k(\xv)\}$. Then the condition above is the same as that $S(\xv,k)$ is a monotonically decreasing function with respect to $k$ for any fixed $\xv\in\Sc$, 
\begin{align}
	S(\xv,1)\geq S(\xv,2)\geq \dots \geq S(\xv,K-1).\label{eq:noncrossing}
\end{align}



\subsection{Direct NORDIC}
The noncrossing condition (\ref{eq:noncrossing}) can be fairly difficult to implement. We consider a sufficient condition first in this subsection. In this article, we use SVM as the basic binary classifier. For a Mercer kernel function $K(\cdot,\cdot)$, the Representer Theorem \citep{kimeldorf1971some} allows the $k$th classification function to be represented by $f_k(\xv)=\sum_{j=1}^n\omega_{k,j}K(\xv_j,\xv)+b_k$.

Note that if we add the constraints that
$$\omega_{k,i}\geq \omega_{k+1,i}~\mb{and}~b_k\geq b_{k+1}~\mb{for}~k=1,\dots,K-2,$$ then as
long as the kernel function is always nonnegative with $K(\cdot,\cdot)\geq 0$ (which is true for many kernel functions such as the Gaussian radial basis function kernel), we will have
$f_k(\xv)\geq f_{k+1}(\xv)$, and hence $S(\xv,k)\geq S(\xv,k+1)$ for any $\xv\in \Sc$.

Hence we consider a direct approach to NORDIC, called NORDIC-0, by solving the following joint optimization problem with the extra noncrossing constraints (\ref{constraint_b})--(\ref{constraint_omega}):
\begin{align}
  \min_{\omega_{k,j},b_k} & \sum_{k=1}^{K-1}\bigg[\sum_{i=1}^n\big(1-y_{i}^{(k)}f_k(\xv)\big)_++\frac{\lambda}{2}\omegav_{k\cdot}^T\Kv\omegav_{k\cdot}\bigg],\label{simpleNORDICformula}
\end{align}
where $f_k(\xv)=\sum_{j=1}^n\omega_{k,j}K(\xv_j,\xv_i)+b_k$, the coefficient vector for the
$k$th function is $\omegav_{k,\cdot}\equiv(\omega_{k,1},\dots,\omega_{k,n})^T$, and $\Kv$ is an $n$ by $n$ matrix whose $(i,j)$th entry is $\Kv_{i,j}=K(\xv_i,\xv_j)$, subject to
\begin{align}
  \quad &b_k\geq b_{k+1},~\mb{for}~k=1,\dots, K-2,\label{constraint_b}\\
  &\omega_{k,i}\geq \omega_{k+1,i},~\mb{for}~i=1,\dots n,~k=1,\dots, K-2.\label{constraint_omega}
\end{align}
Here
$\omegav_{k\cdot}^T\Kv\omegav_{k\cdot}$ is the regularization term for the $k$th discriminant function. 

The term inside the square bracket of (\ref{simpleNORDICformula}) is the objective function of kernel SVM corresponding to the $k$th classifier. We try to minimize the sum of these $(K-1)$ objective functions with the extra noncrossing constraints  (\ref{constraint_b})--(\ref{constraint_omega}).

\subsection{Indirect NORDIC}
The constraints (\ref{constraint_b})--(\ref{constraint_omega}) for NORDIC-0 are sufficient conditions for noncrossing boundaries. However, such condition may be too strong. A weaker, but \textit{almost} sufficient set of conditions would be inequality (\ref{constraint_b}) along with the inequality that
$\Kv\omegav_{k\cdot}\geq \Kv\omegav_{(k+1) \cdot},~\mb{for}~k=1,\dots, K-2.$
Note that they ensure that $f_k(\xv_i)\geq f_{k+1}(\xv_i)$ for all the data $\xv_i$ in the training data set. Thus when the training data is rich enough to cover the base of $\Sc$, then they are \textit{almost} sufficient conditions for noncrossing. This approach is an indirect approach to noncrossing through the training data points, which is called NORDIC-1 in this article. A bonus of this set of constraints compared to (\ref{constraint_b})--(\ref{constraint_omega}) is that one does not need to take the inverse of $\Kv$ later in the implementation, which we will explain in the next subsection.

Let $\yv_{k\cdot}=(y_{1}^{(k)},\dots y_{n}^{(k)})^T$ be the dummy class label vector of the $n$ observations for the $k$th classifier, and $\ev=(1,\dots 1)^T$. For neatness, we let $\Yv_{k\cdot}$ denote the diagonal matrix with $\yv_{k\cdot}$ as its diagonal elements, \ie, $\Yv_{k\cdot}\equiv \diag(\yv_{k\cdot})$. By replacing the Hinge loss $\{1-y_{i}^{(k)}f_k(\xv_i)\}_+$ in (\ref{simpleNORDICformula}) by a slack variable $\xi_{k,i}\geq 0$, and incorporating the new constraints, we can write the optimization problem for NORDIC-1 as,
\begin{align}
  \min_{\omegav_{k\cdot},b_k,\xiv_{k\cdot}}\quad & \sum_{k=1}^{K-1}\left(\frac{1}{2}\omegav_{k\cdot}^T\Kv\omegav_{k\cdot}+C\ev^T \xiv_{k\cdot} \right),\label{primal_problem}
\end{align}subject to
  \begin{align}&\ev-\Yv_{k\cdot}(\Kv\omegav_{k\cdot}+b_k\ev)\leq
						\xiv_{k\cdot}, ~\mb{for}~ k=1,\dots, K-1,\label{primal_constraint1}\\
						&\xiv_{k\cdot}\geq \0v, ~\mb{for}~ k=1,\dots, K-1,\label{primal_constraint2}\\
						&b_k\geq b_{k+1},~\mb{for}~ k=1,\dots, K-2,\label{primal_constraint3}\\
						&\Kv\omegav_{k\cdot}\geq \Kv\omegav_{(k+1) \cdot}, ~\mb{for}~ k=1,\dots, K-2.\label{primal_constraint4}
\end{align}

\subsection{Implementations of NORDIC}
We start off by deriving the Wolfe duality of the optimization problem for NORDIC-1. The implementation of NORDIC-0 will come clearer later as a variant of that of NORDIC-1. We introduce nonnegative Lagrange multipliers
$\alphav_{k\cdot}=(\alpha_{k,1},\dots \alpha_{k,n})^T\in \R_+^n$, $\zetav_{k\cdot}=(\zeta_{k,1},\dots \zeta_{k,n})^T\in \R_+^n$, $\gamma_k\in \R_+$ and
$\varphiv_{k\cdot}=(\varphi_{k,1},\dots \varphi_{k,n})^T\in \R_+^n$ for the constraints (\ref{primal_constraint1}), (\ref{primal_constraint2}), (\ref{primal_constraint3}) and (\ref{primal_constraint4}) respectively. The Lagrangian for the primal problem (\ref{primal_problem})--(\ref{primal_constraint4}) is,
\begin{align*}
\Lc=&\sum_{k=1}^{K-1}\bigg[\left(\frac{1}{2}\omegav_{k\cdot}^T\Kv\omegav_{k\cdot}+C\ev^T
\xiv_{k\cdot}\right)\\
&+ \alphav_{k\cdot}^T\{\ev-\Yv_{k\cdot}(\Kv\omegav_{k\cdot}+b_k\ev)-
\xiv_{k\cdot}\}\\
		&-\zetav_{k\cdot}^T\xiv_{k\cdot}- \gamma_k(b_k-b_{k+1})\Ind{k\neq K-1}\\
		&-\varphiv_{k\cdot}^T(\Kv\omegav_{k\cdot}-\Kv\omegav_{(k+1)\cdot})\Ind{k\neq K-1}\bigg].
\end{align*}
It can be rearranged, so that in the square bracket, the subscripts for the primal variables are with the same index $k$, as follows,
\begin{align}
\Lc=&\sum_{k=1}^{K-1}\bigg[\left(\frac{1}{2}\omegav_{k\cdot}^T\Kv\omegav_{k\cdot}+C\ev^T
\xiv_{k\cdot}\right) \label{Lagrange2}\\
&+ \alphav_{k\cdot}^T\{\ev-\Yv_{k\cdot}(\Kv\omegav_{k\cdot}+b_k\ev)-
\xiv_{k\cdot}\}\notag\\
&		-\zetav_{k\cdot}^T\xiv_{k\cdot}- b_k\left(\gamma_k\Ind{k\neq K-1}-\gamma_{k-1}\Ind{k\neq 1}\right)\notag
\\
&-\omegav_{k\cdot}^T\Kv\left(\varphiv_{k\cdot}\Ind{k\neq K-1}-\varphiv_{(k-1)\cdot}\Ind{k\neq 1}\right)\bigg].\notag
\end{align}
The Karush-Kuhn-Tucker (KKT) conditions for the primal problem require the following:
\begin{align}
\0v=\frac{\partial\Lc}{\partial\omegav_{k\cdot}}&=\Kv\omegav_{k\cdot}- \Kv \Yv_{k\cdot}\alphav_{k\cdot}\label{KKT_omega}\\
&-\Kv\left(\varphiv_{k\cdot}\Ind{k\neq K-1}-\varphiv_{(k-1)\cdot}\Ind{k\neq 1}\right),&\notag\\
0=\frac{\partial\Lc}{\partial b_k}&=-\yv_{k\cdot}^T\alphav_{k\cdot}\label{KKT_b}\\
&-\left(\gamma_k\Ind{k\neq K-1}-\gamma_{k-1}\Ind{k\neq 1}\right),&\notag\\
\0v=\frac{\partial\Lc}{\partial \xiv_{k\cdot}}&=C\ev-\alphav_{k\cdot}-\zetav_{k\cdot}.\label{KKT_xi}&
\end{align}
Once the KKT conditions (\ref{KKT_b}) and (\ref{KKT_xi}) are inserted to (\ref{Lagrange2}), the items that are associated with $b_k$ and $\xiv_{k\cdot}$ will be eliminated. Moreover, from (\ref{KKT_omega}), we have $$\Kv\omegav_{k\cdot}=\Kv\left\{\Yv_{k\cdot}\alphav_{k\cdot}+\left(\varphiv_{k\cdot}\Ind{k\neq K-1}-\varphiv_{(k-1)\cdot}\Ind{k\neq 1}\right)\right\},$$ which leads to 
$$\omegav_{k\cdot}=\Yv_{k\cdot}\alphav_{k\cdot}+\left(\varphiv_{k\cdot}\Ind{k\neq K-1}-\varphiv_{(k-1)\cdot}\Ind{k\neq 1}\right)$$ when $\Kv$ is full rank. Let 
\begin{align*}
	\Rv &= \Big[\diag\set{\Yv_{k\cdot}}_{1\leq k\leq K-1}\mid \Id_{n(K-1)}^{(n)}\\
	&\quad\quad\quad-\Id_{n(K-1)}^{(-n)}\mid\0v_{n(K-1)\times (K-2)}\Big]
\end{align*}
 and $\thetav = (\alphav; \varphiv;\gammav)$, where for a $m\times n$ matrix $\Av$, $\Av^{(s)}$ denotes a $(m+s)\times n$ matrix whose upper $m$ rows are $\Av$ and the bottom $s$ rows are all 0, and $\Av^{(-s)}$ denotes a $(m+s)\times n$ matrix whose bottom $m$ rows are $\Av$ and the top $s$ rows are all 0. Summarizing all these conditions, we can see that the optimality of the primal problem is given by the dual problem, 
\begin{align*}
\max_{\thetav_k \equiv (\alphav_{k\cdot}; \varphiv_{k\cdot};\gamma_k)} &-\frac{1}{2}\thetav^T\left\{\Rv^T\left(\Id_{K-1}\otimes \Kv\right)\Rv\right\}\thetav+\ev^T\alphav,\\
  \mb{subject to}\quad & -\yv_{k\cdot}^T\alphav_{k\cdot}-\left(\gamma_k\Ind{k\neq K-1}-\gamma_{k-1}\Ind{k\neq 1}\right)=0,\\
  &\0v\leq\alphav_{k\cdot}\leq C\ev,~\varphiv_{k\cdot}\geq\0v,~\gamma_k\geq 0,
\end{align*}where $\otimes$ is the Kronecker product.

The dual problem above is nothing but a quadratic programming (QP) problem about $\alphav_{k\cdot}, \varphiv_{k\cdot}, \gamma_k$ with equality  and bound-inequality constraints, which can be solved by many third-party off-the-shelf QP subroutines. More efficient implementations, such as Platt's SMO \citep{Platt1999Fast}, are possible, but is not explored here as it is beyond the scope of this paper.

The optimal primal variables $\omegav$ are calculated from  the optimal dual variables using the relation $\omegav_{k\cdot}=\Yv_{k\cdot}\alphav_{k\cdot}+\left(\varphiv_{k\cdot}\Ind{k\neq K-1}-\varphiv_{(k-1)\cdot}\Ind{k\neq 1}\right)$. By the KKT complementary conditions, the bias term $b_k$ for the $k$th classifier can be found from any $\xv_i$ in the training data with $0 \leq \alpha_{k,i}\leq C$, due to the relations that $1-y_{i}^{(k)}\{\sum_{j=1}^{n}\omega_{kj}K(\xv_j,\xv_i)+b_k\}=0$. Alternatively, one can fix the $\omegav$'s in the primal (\ref{primal_problem}) as known and minimize (\ref{primal_problem})--(\ref{primal_constraint4}) with respect to $b_k$ and $\xiv_{k\cdot}$. This would lead to a linear programming problem.

The implementation for NORDIC-0 is similar, except that the Lagrangian is
\begin{align*}
\Lc_0=&\sum_{k=1}^{K-1}\bigg[\left(\frac{1}{2}\omegav_{k\cdot}^T\Kv\omegav_{k\cdot}+C\ev^T
\xiv_{k\cdot}\right)\\
&+ \alphav_{k\cdot}^T\{\ev-\Yv_{k\cdot}(\Kv\omegav_{k\cdot}+b_k\ev)-
\xiv_{k\cdot}\}\\
		&-\zetav_{k\cdot}^T\xiv_{k\cdot}- \gamma_k(b_k-b_{k+1})\Ind{k\neq K-1}\\
		&-\varphiv_{k\cdot}^T(\underline{\omegav_{k\cdot}-\omegav_{(k+1)\cdot}})\Ind{k\neq K-1}\bigg].
\end{align*}
The only difference of the Lagrangian of NORDIC-0 from that of NORDIC-1 is underlined. Consequently, the KKT conditions are almost the same, except that,
\begin{align*}
\0v=\frac{\partial\Lc_0}{\partial\omegav_{k\cdot}}&=\Kv\omegav_{k\cdot}- \Kv \Yv_{k\cdot}\alphav_{k\cdot}\\
&-\left(\varphiv_{k\cdot}\Ind{k\neq K-1}-\varphiv_{(k-1)\cdot}\Ind{k\neq 1}\right).&
\end{align*}
This leads to $\omegav_{k\cdot}$ at the optimality being $$\omegav_{k\cdot}=\Yv_{k\cdot}\alphav_{k\cdot}+\Kv^{-1}\left(\varphiv_{k\cdot}\Ind{k\neq K-1}-\varphiv_{(k-1)\cdot}\Ind{k\neq 1}\right),$$ assuming that $\Kv$ is invertible. The rest of the implementation is identical to that in NORDIC-0, except that we let 
\begin{align*}
	\Rv = \Big[\diag\set{\Yv_{k\cdot}}_{1\leq k\leq K-1}\mid \{\Id_{K-1}\otimes\Kv^{-1}\}^{(n)}\\-\{\Id_{K-1}\otimes\Kv^{-1}\}^{(-n)}\mid\0v_{n(K-1)\times (K-2)}\Big]
\end{align*} and $\thetav = (\alphav; \varphiv;\gammav)$.

\section{Exact NORDIC via Integer Programming}\label{sec:NORDICIP}
Recall that the necessary and sufficient condition for noncrossing (\ref{eq:noncrossing}) is that the sign of $f_k(\xv)$, $S(\xv,k)$, is a monotonically decreasing function with respect to $k$ for any fixed $\xv\in\Sc$, $S(\xv,1)\geq S(\xv,2)\geq \dots \geq S(\xv,K-1).$ The constraints for NORDIC-0 and NORDIC-1 that we have discussed in the last section is sufficient to ensure that 
$f_1(\xv)\geq f_2(\xv)\geq \dots \geq f_{K-1}(\xv),$ which ultimately ensures noncrossing. However, they are not the weakest sufficient conditions we can impose. As a matter of fact, the discriminative functions $f_k$ themselves need not to be monotonically decreasing with respect to $k$ in order for noncrossing. In this section, we explore an idea which aims for exact noncrossing by posing conditions on the sign of the discriminative functions.

For each $\xv\in\Sc$, there are one out of two alternative situations with regard to the prediction result from a discriminant function $f_k$: either $f_k(\xv)< 0$ or $f_k(\xv)\geq 0$. According to the noncrossing condition (\ref{eq:noncrossing}), the former implies that $f_{k+1}(\xv)<0$ (recall that the sign is monotonically decreasing in $k$). Thus, the noncrossing condition (\ref{eq:noncrossing}) is logically equivalent to the condition that at least one of the following two constraints is satisfied,
$$\mb{(i)}~f_k(\xv)\geq 0,~\mb{and}~\mb{(ii)}~f_{k+1}(\xv)\leq 0;$$ \ie, (i) and (ii) cannot be both false.
Specifically, if (i) is not true, \ie, if $f_k(\xv)< 0$, then (ii) is true. This leads to the noncrossing condition.

Such logical implication can be modeled by the following \textit{Logical Constraints} which involve binary integer variables $z_{1k},~z_{2k}\in\set{0,1}$,
\begin{align*}
	-f_k(\xv) - M_1z_{1k}&\leq 0, \\
	f_{k+1}(\xv) - M_2z_{2k}&\leq 0, \\
	z_{1k}+z_{2k}&\leq 1,
\end{align*}
where $M_1$ and $M_2$ are two large numbers due to technicality. In particular, $z_{1k}+z_{2k}\leq 1$ implies that at least one between $z_{1k}$ and $z_{2k}$ has to be zero, hence (considering the first two constraints) either $-f_k(\xv)\le 0$ or $f_{k+1}(\xv)\le 0$, or both are true; this is the noncrossing condition discussed above. Note that if both $z_{1k}$ and $z_{2k}$ were 1, then the first two constraints became $-f_k(\xv)\le M_1$ and $f_{k+1}(\xv)\le M_2$, which would essentially impose no constraint on $f_k(\xv)$ and $f_{k+1}(\xv)$ so that the undesired case that $f_k(\xv)<0$ and $f_{k+1}(\xv)>0$ may occur. See \citet{bradley1977applied} for an introduction to integer programming. We can use this technique to model the noncrossing constraints. In particular, we seek to
\begin{align}
  \min_{\omega_{k,j},b_k} & \sum_{k=1}^{K-1}\left[\sum_{i=1}^n\left(1-y_{i}^{(k)}f_k(\xv)\right)_++\lambda\|\omegav_{k\cdot}\|_1\right],
\end{align}
subject to
\begin{align}
  &-f_k(\xv_i) - M_1z_{1ik}\leq 0,\label{cond1}\\
	&f_{k+1}(\xv_i) - M_2z_{2ik}\leq 0,\\
	&z_{1ik}+z_{2ik}\leq 1,\\
	&z_{1ik},~z_{2ik}\in\set{0,1},\label{cond2}
\end{align}
for $i=1,2,\dots,n~\mb{and}~k=1,\dots, K-2.$
This method is referred to as NORDIC-2 in this article. Here the constrains (\ref{cond1})--(\ref{cond2}) are almost sufficient and (exactly) necessary conditions to \textit{noncrossing}. It is again not exactly sufficient because we impose the constraints to all the training data vectors, instead of all $\xv\in\Sc$, similar to the case of NORDIC-1. However, again, if the data vectors in the training data are rich enough, noncrossing across the board can be expected. These conditions are weaker than those in NORDIC-0 and NORDIC-1 because they ensure the monotonicity of the sign of $f_k$, rather than the value of $f_k$ itself.

Note that the objective function of NORDIC-2 is a little different from those of NORDIC-0 and NORDIC-1, especially in the use of the $L_1$ norm penalty. We choose not to use the more common $L_2$ penalty, which leads to a quadratic objective function in SVM, because it is rather difficult to solve a mixed integer programing problem with quadratic objective function. In fact, we are not aware of an efficient off-the-shelf computing freeware which solves such a problem. In order to show the usefulness of the new noncrossing constraints, which is the main point of this article, we choose to use the $L_1$ penalty for computational simplicity.

It is worth noting that so long as there is an efficient mixed integer programming package which is capable of dealing with quadratic objective functions, an extension will be very natural and readily available.

Indeed, integer programming can solve such nonstandard problem which traditional optimization methods such as QP or linear programming cannot. However, integer programming can been overlooked by statisticians for a long time (probably due to the high computational cost and few statistical problem that this method applies). To the author's best knowledge, this article is one of only a few work in the statistical literature which employs the integer programming technique. See \citet{Liu2006} for another instance which uses mixed integer programming to solve a statistical problem.

\section{Theoretical Properties}\label{sec:theory}
In this section, we study two aspects of the theoretical properties of NORDIC. The first subsection is about the Bayes rule and Fisher consistency of the loss function in ordinal classification. The second  one pertains to the asymptotic normality of the NORDIC solution.
\subsection{Bayes rules and Fisher Consistency}
For binary classification, a classifier with loss $V_1(yf(\xv)):\R\mapsto\R_+$ is Fisher consistent if the minimizer of $\E[V_1(Yf(\Xv))|\Xv=\xv]$  has the same sign as $\Pr(Y=1|\Xv=\xv)-1/2$. The latter is the Bayes rule for binary classification. Intuitively, Fisher consistency requires that the classifier yields the Bayes decision rule asymptotically. See \citet{lin2004note} for Fisher consistency of binary large margin classifiers.

In multicategory classification, a classifier with loss function $V_2(y,\fv(\xv)):\R\times\R^K\mapsto\R_+$, where $\fv(\xv):\Sc\mapsto\R^K$ is the $K$ discriminant functions, is Fisher consistent if the minimizer of $\E[V_2(Y,f(\Xv))|\Xv=\xv]$, $\gv^*(\xv)=(g_1^*(\xv),\dots,g_K^*(\xv))^T$, satisfies that $\argmax_{k\in\set{1,\dots,K}}g_k^*(\xv)=\argmax_{k\in\set{1,\dots,K}}\eta_k(\xv)$. Here, $\argmax_{k\in\set{1,\dots,K}}\eta_k(\xv)$ is the Bayes classification rule for multicategory classification. See, for example, \citet{Liu2007} for some discussions on Fisher consistency for multicategory SVM classifiers.

Below we formally define the Bayes rule and Fisher consistency for ordinal classification. The Bayes rule for ordinal classification is 
$\phi^{Bayes}_{OC}(\xv)=k$ where $k$ is such that $\sum_{\ell=1}^{k-1} \eta_{\ell}(\xv)<1/2$ and $\sum_{\ell=1}^{k}\eta_{\ell}(\xv)>1/2$. This rule guarantees that each component binary classification in ordinal classification yields the Bayes rule in the binary sense.

\begin{definition}\textit{(Generalized Fisher consistency for ordinal classification)} An ordinal classification method with loss function $V_3(\cdot,\cdot)$ is Generalized Fisher consistent if for any $\xv$, the minimizer $\fv^*(\xv)=(f_1^*(\xv),\dots,f_{K-1}^*(\xv))^T$ of $$\E\left[\sum_{k=1}^{K-1}V_3(Y^{(k)},f_k(\Xv))\right]$$ satisfies that $\sgn(f_k^*(\xv)) = \sgn(1/2-\sum_{\ell=1}^k\eta_\ell(\xv))$ for $k=1,\dots,K-1$. Here $Y^{(k)}$ is the dummy class label for Class $Y$ in the $k$th binary classification subproblem.
\end{definition}

Generalized Fisher consistency means that the $(K-1)$ discriminant functions $f_1^*(\xv),\dots,f_{K-1}^*(\xv)$ jointly trained under the loss function $V_3$, is essentially the same as the Bayes rule $\phi^{Bayes}_{OC}(\xv)$, as $n\rightarrow\infty.$ Note that $\phi_{OC}^{Bayes}(\xv)$ has the smallest risk with respect to the aggregated 0-1 loss for the $(K-1)$ binary subproblems. Hence it is also the one which has the smallest risk under the so-called distance loss, defined as $L(\phi,y)=|\phi-y|$ \citep[see][]{Qiao2015Learning}.

Because of the use of the Hinge loss function for SVM (which is Fisher consistent in the binary sense), our NORDIC method is Generalized Fisher consistent for ordinal classification. The proof is omitted.

\subsection{Asymptotic Normality of Linear NORDIC}
When the kernel function $K(\xv_1,\xv_2)=\xv_1^T\xv_2$, that is, the linear kernel, we can have the following linear NORDIC classifier, with the objective function,
\begin{align}
  \sum_{k=1}^{K-1}\left[\sum_{i=1}^n\left(1-y_{i}^{(k)}\left(\xv_i^T\omegav_{k\cdot}+b_k\right)\right)_++\frac{\lambda}{2}\omegav_{k\cdot}^T\omegav_{k\cdot}\right],\label{linearNORDIC}
\end{align}
and one of the two following sets of constraints that correspond to NORDIC-1 and NORDIC-2 respectively,
\begin{align*}
  \quad &\xv_i^T\omegav_{k\cdot}+b_k\geq \xv_i^T\omegav_{k+1,\cdot}+b_{k+1},
\end{align*}
and
\begin{align*}
	  \quad &	-\left(\xv_i^T\omegav_{k\cdot}+b_k\right) - M_1z_{1k}\leq 0,\\
	&\left(\xv_i^T\omegav_{k+1,\cdot}+b_{k+1}\right) - M_2z_{2k}\leq 0,\\
	&z_{1k}+z_{2k}\leq 1,\\
	&z_{1k},~z_{2k}\in\set{0,1},
\end{align*}for $i=1,2,\dots,n~\mb{and}~k=1,\dots, K-2$.

Because linear kernel could be negative, the NORDIC-0 method cannot be directly extended to the linear kernel case. We can use the technique in \citet{liu2011simultaneous} to create a new kernel that satisfies the nonnegativity assumption essential for NORDIC-0. In this subsection, we prove the asymptotic normality of linear NORDIC.

\citet{koo2008bahadur} has provided a Bahadur representation of the linear SVM and proved its asymptotic normality under some conditions. In particular, they have shown that $(\tilde\omegav,\tilde b)^T - (\omegav^0,b^0)^T=O_p(n^{-1/2})$, where $(\tilde\omegav,\tilde b)$ are the solution to the SVM classifier and $(\omegav^0,b^0)$ are the minimizer of the expected loss function.

Theorem \ref{thm1} below shows that the limiting distribution of the constrained NORDIC solution has the same limiting distribution to the unconstrained binary SVM classifiers. To prove this result, we need all the regularity conditions in \citet{koo2008bahadur}.
\begin{theorem}\label{thm1}
For $1\leq k\leq K-1$, let $(\hat\omegav_{k\cdot},\hat b_k)$ and $(\tilde\omegav_{k\cdot},\tilde b_k)$ be the constrained and unconstrained solutions, respectively, to the $k$th binary linear SVM problem in (\ref{linearNORDIC}). Assume that the regularity conditions in \citet{koo2008bahadur} are satisfied for $k$. Then for any $\uv\in\R^{(d+1)(K-1)}$,
\begin{align*}
	&\Bigg|\Pr\left[n^{1/2}\left\{(\hat\omegav_{k\cdot},\hat b_k)^T-(\omegav_{k\cdot}^0,b_k^0)^T\right\}\leq \uv\right]\\&\quad\quad-\Pr\left[n^{1/2}\left\{(\tilde\omegav_{k\cdot},\tilde b_k)^T-(\omegav_{k\cdot}^0,b_k^0)^T\right\}\leq \uv\right] \Bigg|\rightarrow 0,
\end{align*} so that the constrained solution has the same limiting distribution as the classical unconstrained solution.
\end{theorem}

Based on Theorem \ref{thm1}, inference for the constrained NORDIC can be obtained by applying the known asymptotic results for binary linear SVM, through the unconstrained NORDIC solutions. For example, we can show the asymptotic normality of the coefficients to the SVM components in linear NORDIC in the same way as those in \citet{koo2008bahadur}.

\section{Numerical Results}\label{sec:simulation}
We compare NORDIC-0, NORDIC-1, NORDIC-2, the vanilla ordinal classification method that uses $(K-1)$ separately trained \citep{Frank2001Simple} using binary SVM classifiers  (BSVM), the data replication method by \citet{cardoso2007learning} (DR) and the parallel discriminant hyperplane method by \citet{chu2005new} (CK). We use our own experimental codes in the R environment to implement these methods. The Gaussian radial basis function kernel is used for all classifiers. The kernel parameter is tuned among the 10\%, 50\% and 90\% quantiles of the pairwise distances between training vectors. The tuning parameters are tuned from a grid of possible values ranging from $2^{-4},2^{-3},\dots,2^{4}$.

\subsection{Nonlinear Three-class Examples}\label{sec:3class}
\begin{figure}[!b]
\begin{center}
		\includegraphics[width = 0.6\linewidth]{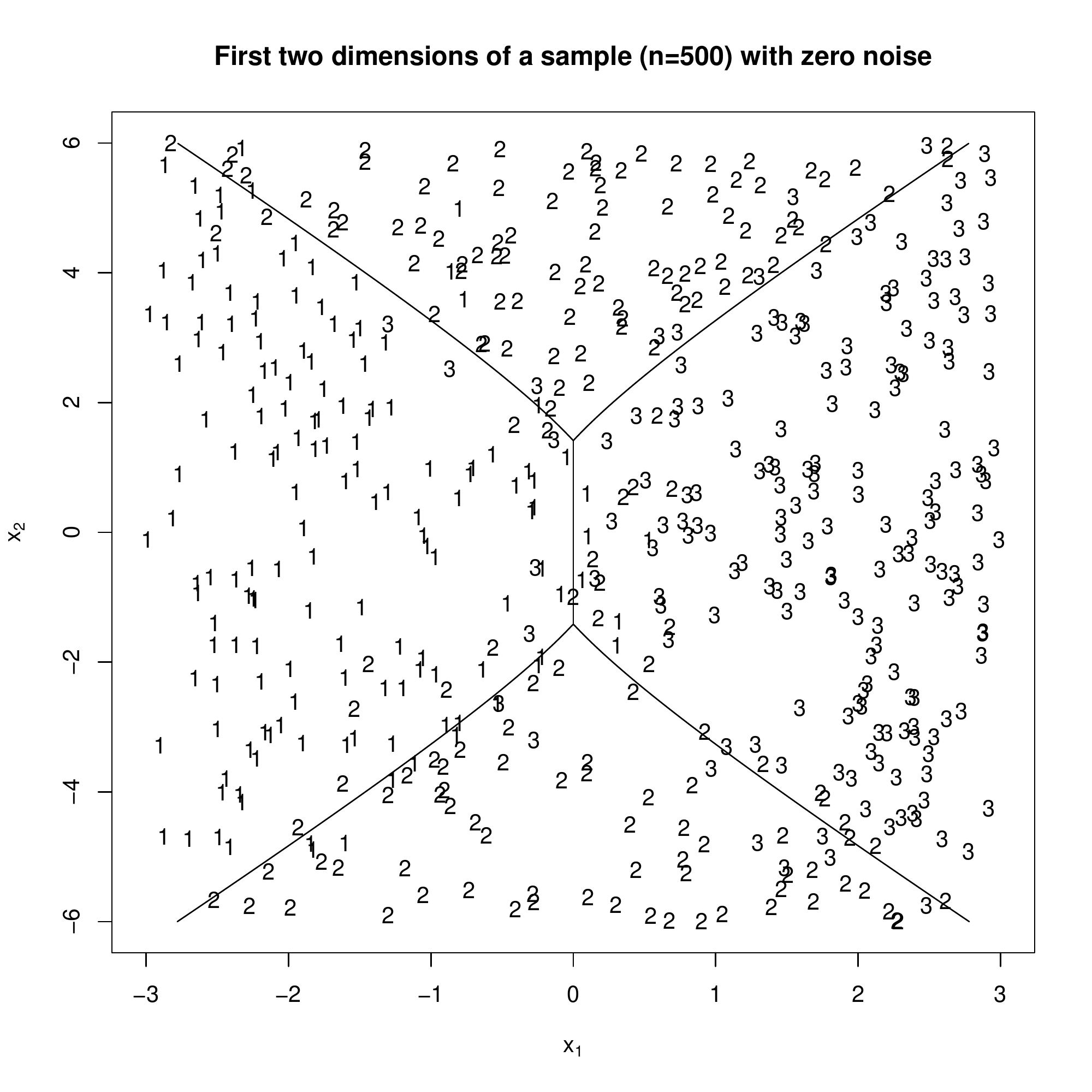}\vspace{-2em}
\end{center}
		\caption{\small Nonlinear three-class examples: A scatter plot showing the first two dimensions of a realization with no additional error added. The Bayes rule is also shown.}
	\label{fig:nonlinear_scatter}
\end{figure}
We consider a data setting with three classes and $d$ variables: $X_1,~X_2,\dots,~ X_d$, where 
\begin{itemize}
	\item $X_1=\tilde X_1+\sigma N(0,1)$ and $\tilde X_1 \sim \mb{Uniform}(-3,3)$, 
	\item $X_2=\tilde X_2+\sigma N(0,1)$ and $\tilde X_2 \sim \mb{Uniform}(-6,6)$,
	\item and $X_3,\dots,X_d \sim N(0,1)$.
\end{itemize} Here, $\tilde X_1$ and $\tilde X_2$ truly determine the class labels (see below) but only their contaminated counterparts $X_1$ and $X_2$ are observed. In particular, let 	
\begin{align*}
  f_1 &= -2\tilde X_1 + 0.2\tilde X_1^2 - 0.1\tilde X_2^2 + 0.2,\\
	f_2 &= -0.4\tilde X_1^2 + 0.2\tilde X_2^2 -0.4,\\
	f_3 &= 2\tilde X_1 + 0.2\tilde X_1^2 -0.1\tilde X_2^2 + 0.2.
\end{align*}
We assign each observation to class $k$ with probability proportional to $\exp(f_k)$ for $k=1,2,3$. We generate 100 data points in the training set, 100 in the tuning set and 10000 in the test set. The standard deviation of the measurement error, $\sigma$, ranges from 0.5, 1 to 1.5. When $d=5$ and $\sigma=0$ (no perturbation), this is the same example as the nonlinear example in \citet{zhang2008variable}. However, we perturb the data and increase the dimension ($d=10,20,\dots,50$) to make the problem more challenging.

Note that this example was initially designed by \citet{zhang2008variable} as a regular multicategory classification, instead of an ordinal classification one. Figure \ref{fig:nonlinear_scatter} shows a sample realization of the data without perturbation at the first two dimensions. In a general sense, Class 2 can be viewed as in the middle of Class 1 and Class 3.  We pretend that the class labels are of an ordinal nature and compare different ordinal classification methods.

\begin{figure}[!t]
\begin{center}
		\includegraphics[width = 0.7\linewidth]{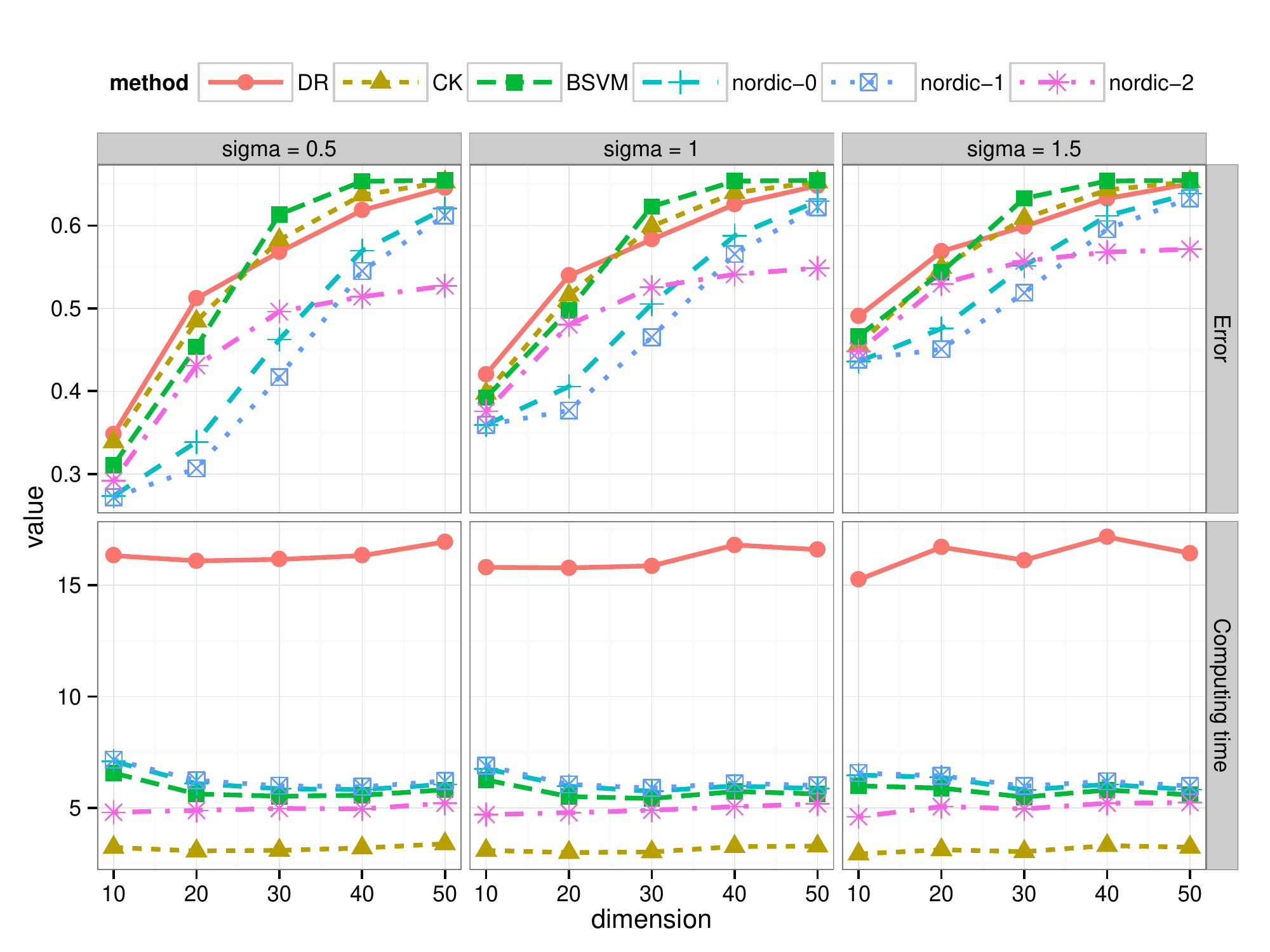}		\vspace{-2em}
\end{center}
		\caption{\small Nonlinear three-class examples: The top row shows the error rate for different methods (in different line types) with 3 noise levels (the left, middle and right panels) and 5 different dimensions (shown on the horizontal axis of each subfigure). The bottom row shows the computational time. In general, a NORDIC method is better than a non-NORDIC method for this example.}
	\label{fig:nonlinear3cls}
\end{figure}

Figure \ref{fig:nonlinear3cls} summarizes the results over 100 simulations. The NORDIC-0 and NORDIC-1 are the better classifiers in terms of classification performance when the dimensions are small. For higher dimensions, the NORDIC-2 method is better than the other methods. 
 The DR method is the most computational costly and the CK method is the most efficient one. The reason that NORDIC works here is probably due to the perturbation that is added to this data set. A NORDIC method, with the help of the noncrossing constraints, can borrow strength from different classes and become more robust to perturbation.

\subsection{Donut Examples}
\begin{figure}[!t]
\centering
		\includegraphics[width = 0.5\linewidth]{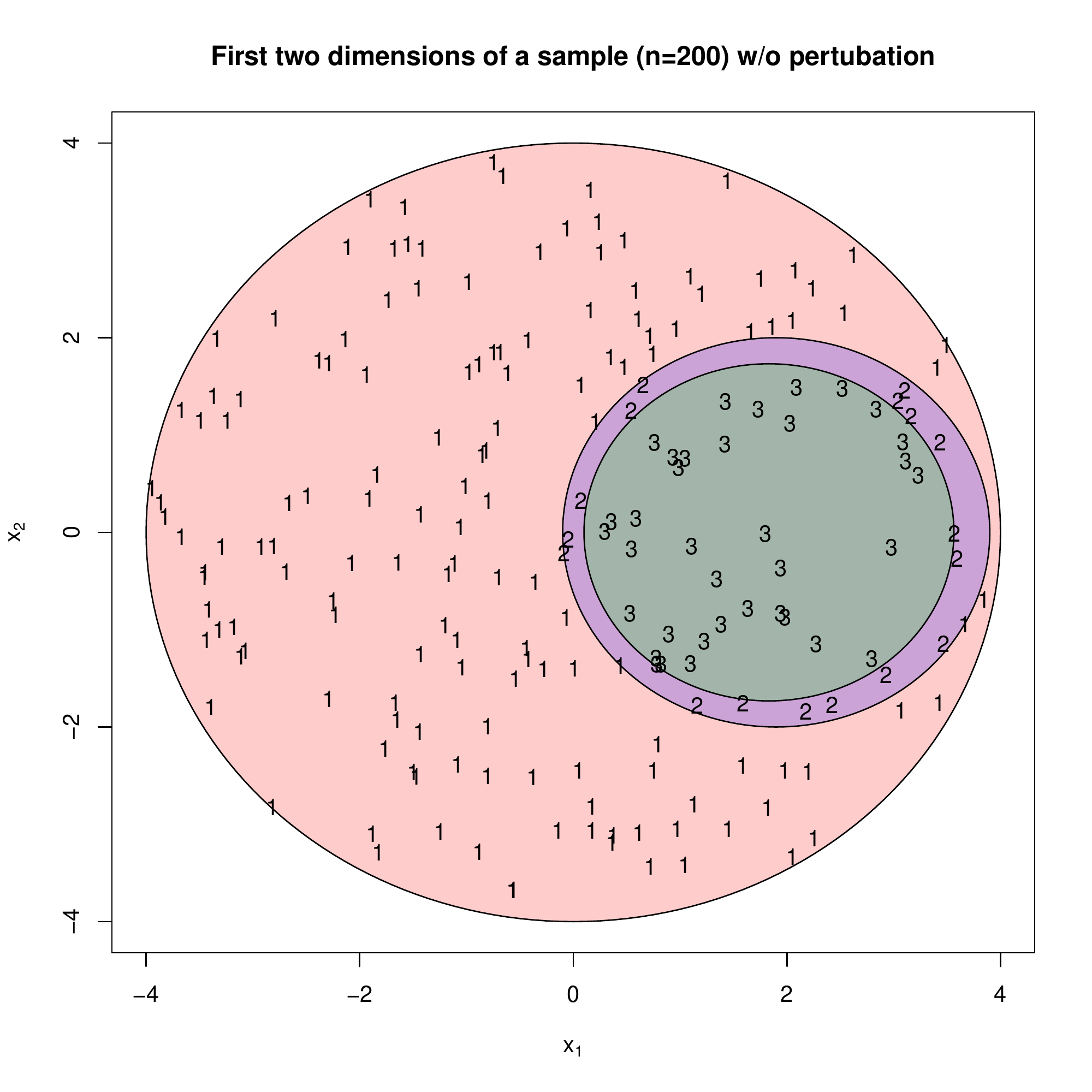}\vspace{-2em}
		\caption{\small Donut examples: A scatter plot showing the first two dimensions of a realization, with no perturbation added. The natural boundaries between classes are also shown.}
	\label{fig:HD_donut_sc}
\end{figure}

We now consider a more challenging setting, which is tailered toward the ordinal data. We first generate data points from a 2-dimensional plate with radius 4 uniformly, and label them as from Class 1, except for those which are within a circle centered at $(1.9,0)^T$ with radius $2$, which are labeled as Class 2, and those which are within a circle centered at $(\sqrt 3+0.1,0)^T$ with radius $\sqrt 3$, which are labeled as Class 3. The observations for the additional $(d-2)$ dimensions are all 0. We then perturb all the data points by adding independent $d$-dimensional Gaussian distributed random vector from $N_d(\0v,\sigma\Id)$. We let $\sigma=0.2$, $0.4$ and $0.6$ and let $d$ range from 10 to 75. Figure \ref{fig:HD_donut_sc} shows one realization of the data on the first two dimensions without the perturbation and the natural boundaries between the classes. This generalizes the classic donut examples in nonlinear classification.

\begin{figure}[!b]
	\centering
		\includegraphics[width = 0.7\linewidth]{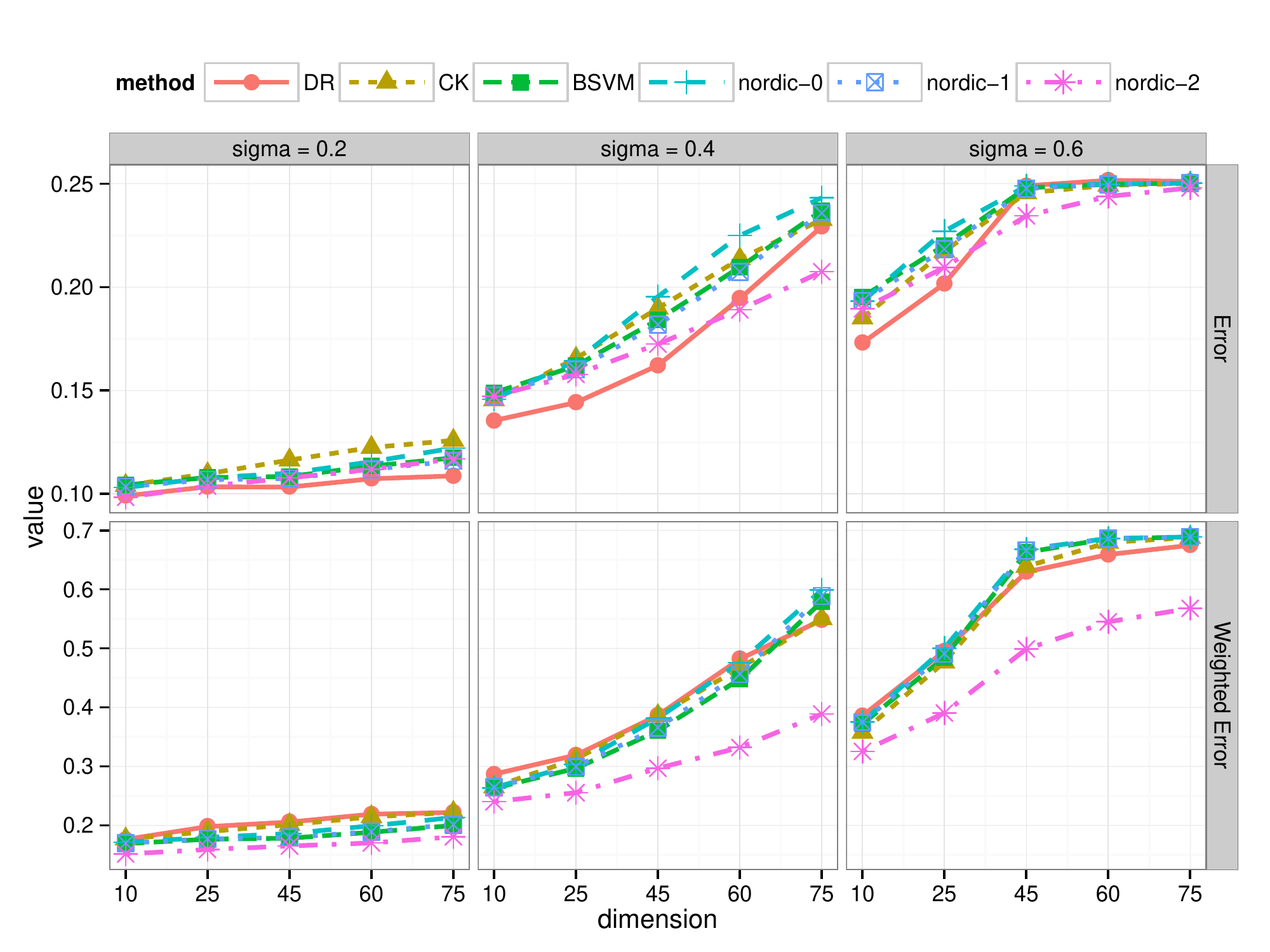}\vspace{-1.5em}
		\caption{\small HD donut examples: The top row shows the error rate for different methods (in different line types) over 12 experiments with 3 noise levels (the left, middle and right panels) and 5 different dimensions (shown on the horizontal axis of each subfigure). The bottom row shows the weighted error rate. The NORDIC-2 is the best classifiers in terms of classification performance.}
	\label{fig:HD_donut}
\end{figure}

Note that Class 2 is sandwiched by Class 1 and Class 3 from both outside and inside, and the high density region for Class 2 is very thin due to the construction. Hence, it is perceivable that a Class 2 observation is very difficult to be correctly classified. The noncrossing constraints here may be of some help because the boundary between Classes 1 and 2 may boost the estimation of the boundary between Classes 2 and 3, and vice versa.

The simulation results are reported in Figure \ref{fig:HD_donut}. The first row shows the test error over 100 simulations. It appears that many times the DR method is the best. However, recall that in this data set the three classes are highly imbalanced in terms of their sample size. On average, there are only 6.25\% Class 2 points and 18.75\% Class 3 points. A more reasonable measure to look into here is some weighted error rate that incorporates the different costs of misclassification. Here we report the weighted error with the configuration that:
\begin{itemize}
	\item each misclassified point from Class 1 costs $1$;
	\item each misclassified point from Class 2 to either Class 1 or Class 3 costs $2$;
	\item each misclassified point from Class 3 to Class 2 costs $1$, and from Class 3 to Class 1 costs $3$.
\end{itemize}
Such assignment of the cost reflects the protection for Class 2, and the additional penalization for misclassifying across two boundaries (the cost for misclassifying from Class 3 to Class 1 is the sum of the costs for misclassifying from 3 to 2 and from 2 to 1.)

\begin{figure}[!b]
\centering
		\includegraphics[width = 0.7\linewidth]{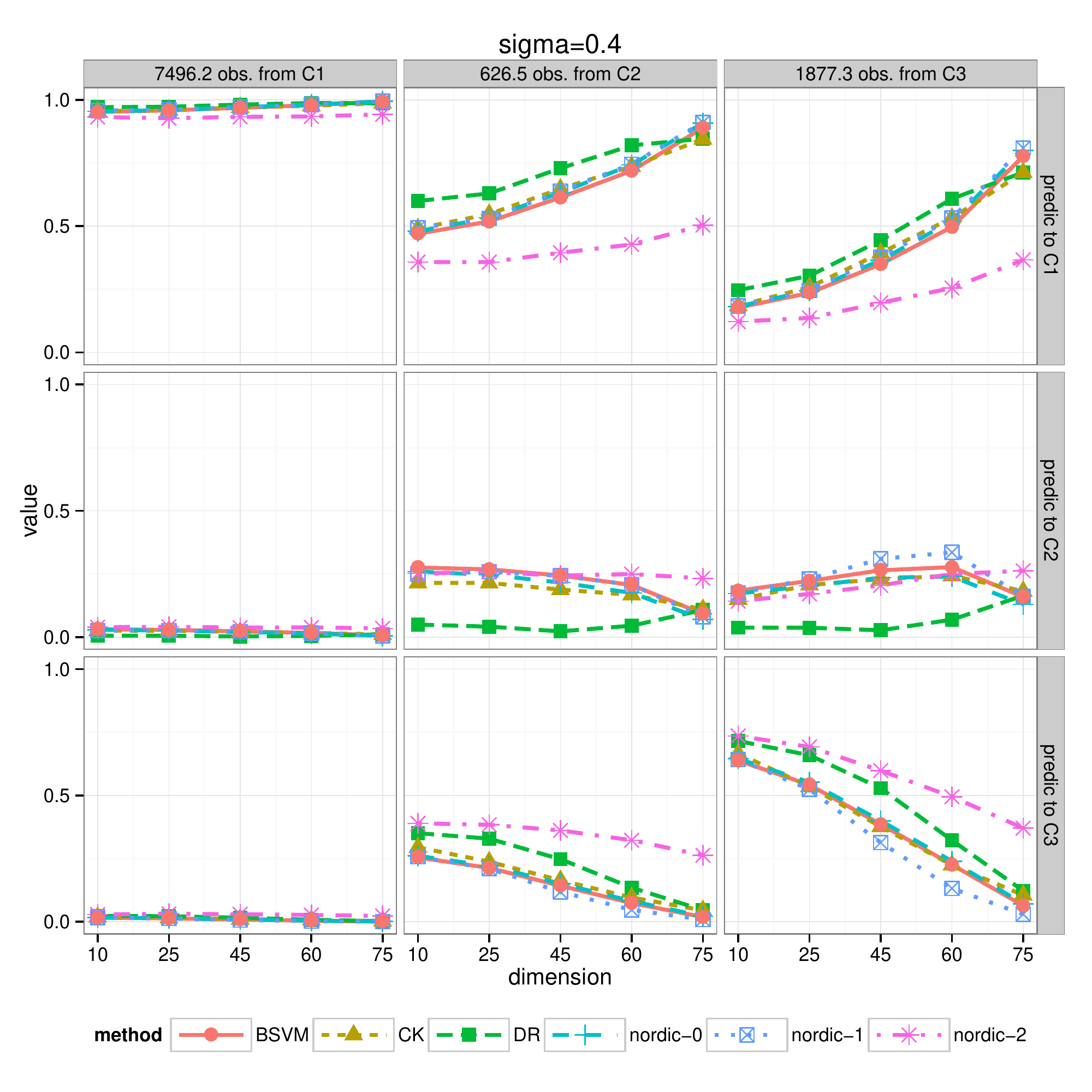}\vspace{-2em}
		\caption{\small Confusion matrices for the examples with $\sigma=0.4$ for different methods and dimensions.}\vspace{-2em}
	\label{fig:newdonut_rev_noise1}
\end{figure}

The second row of Figure \ref{fig:HD_donut} reports the weighted error rate. It is obvious that expect for NORDIC-2, which is the best in this case, all other methods are more or less the same in terms of the weighted error. Interestingly, the NORDIC-0 and NORDIC-1 methods do not perform as well as their sibling NORDIC-2. They perform comparably to the other methods (they may have a very small advantage over CK and DR methods when the perturbation is small, for example, when $\sigma=0.2$ and $0.4$.) Recall that NORDIC-0 and NORDIC-1 imposes stronger constraints which aim for the monotonicity of the discriminant function $f_k(\xv)$ itself, as opposed to its sign. In contrast, the constraint from NORDIC-2 is much lighter, which may have left enough ``degrees of freedom'' to optimize the generalization performance.

One may argue that the choice of the costs in the weighted error may be arbitrary. In this case, it may be helpful to look into the confusion matrix to see the cause of the different performance. Figure \ref{fig:newdonut_rev_noise1} depicts the $3\times 3$ confusion matrices for the case with contamination $\sigma=0.4$ for different methods and different dimensions. For the $(k,\ell)$th plot in the array, the reported value is the proportion of observations from Class $\ell$ that are classified to Class $k$ ($\ell,k=1,2,3$). Note that the aggregation of the three plots in the same column equals to 1. A good classifier is expected to have high rates in the diagonal plots and low rates in the off-diagonal plots. There are, on average, 7496.2 observations from Class 1, and almost all the methods classify them correctly. Class 2 (with only 626.5 observations) is clearly a very difficult class. Even our NORDIC-2 has a poor classification accuracy of $25\%$. That said, NORDIC-2 shows more advantages for higher dimensional cases. For Class 3, NORDIC-2 shows improved accuracy, especially with much fewer misclassifications into Class 1 (shown in the upper-left plot).

The computational time results are similar to what we have seen for the last example and are not reported here.

\section{Real Application}\label{sec:real}
We use the scale balance data set from the UCI Machine Learning Repository \citep{Lichman2013UCI} to test the usefulness of the NORDIC method. This data set, studied in \citet{Siegler1976Three},  was generated to model psychological experimental results. Each example is classified as having the balance scale tip to the right, tip to the left, or be balanced. The four attributes are the left weight, the left distance, the right weight, and the right distance. The correct way to find the class is the greater of (left-distance * left-weight) and (right-distance * right-weight). If they are equal, it is balanced. There are 625 instances in the data, with 288 tip to the left ($L$), 49 balanced ($B$),  and 288 tip to the right ($R$).

\begin{figure}[!t]
	\centering
		\includegraphics[width =0.7\linewidth]{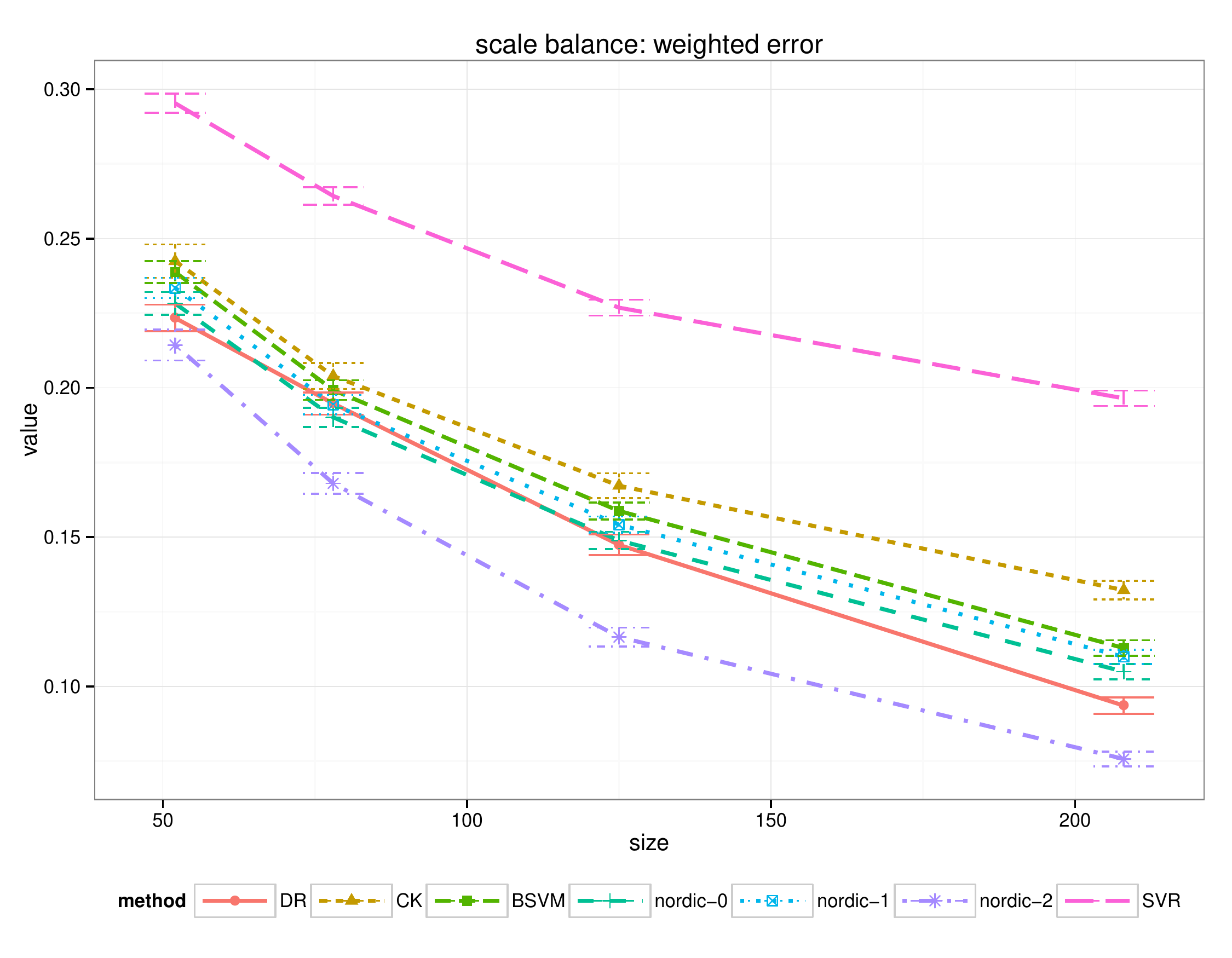}\vspace{-2em}
		\caption{\small Weighted error rate for the scale balance data set.}
	\label{fig:scale}
\end{figure}

There is a clear order between the three classes ($L$, $B$ and $R$,) and hence ordinal classification methods are appropriate. We randomly select $n$ points from the data set for training, $n$ for tuning, and the remaining $(625-2n)$ are for testing. The proportions of the three classes are preserved when the partitioning is conducted. The random experiment is repeated for 100 times.  We consider four cases, where $n=52,~79,~125$ and $208$ respectively.

A naive coding of 1, 2 and 3 for these three classes followed by a regression method will prove to be suboptimal. In particular, in addition to the ordinal classification methods, we also compare with support vector regression \citep[SVR][implemented by \texttt{svm()} in the  R package \texttt{e1071}]{Smola2004tutorial} with Gaussian radial basis function kernel. SVR is applied to the data with \{1,2,3\} coding, and the predicted class is obtained by cut-off values $1.5$ and $2.5$ for the predicted numerical outcome.

Figure \ref{fig:scale} shows the weighted error rates of different methods over 100 random splitting of the data set and 4 different sample sizes. Here we let a misclassified point from Class 3 to Class 1, or from Class 1 to Class 3, to bear a cost of $2$; other types of misclassification cost only $1$. All three NORDIC methods are among the best, with NORDIC-2 having a significant advantage. The other two NORDIC methods are comparable to the DR method especially for small sample cases. The SVR is the worst classifier in this experiment.

\begin{figure}[b!]
	\centering
		\includegraphics[width=0.7\linewidth]{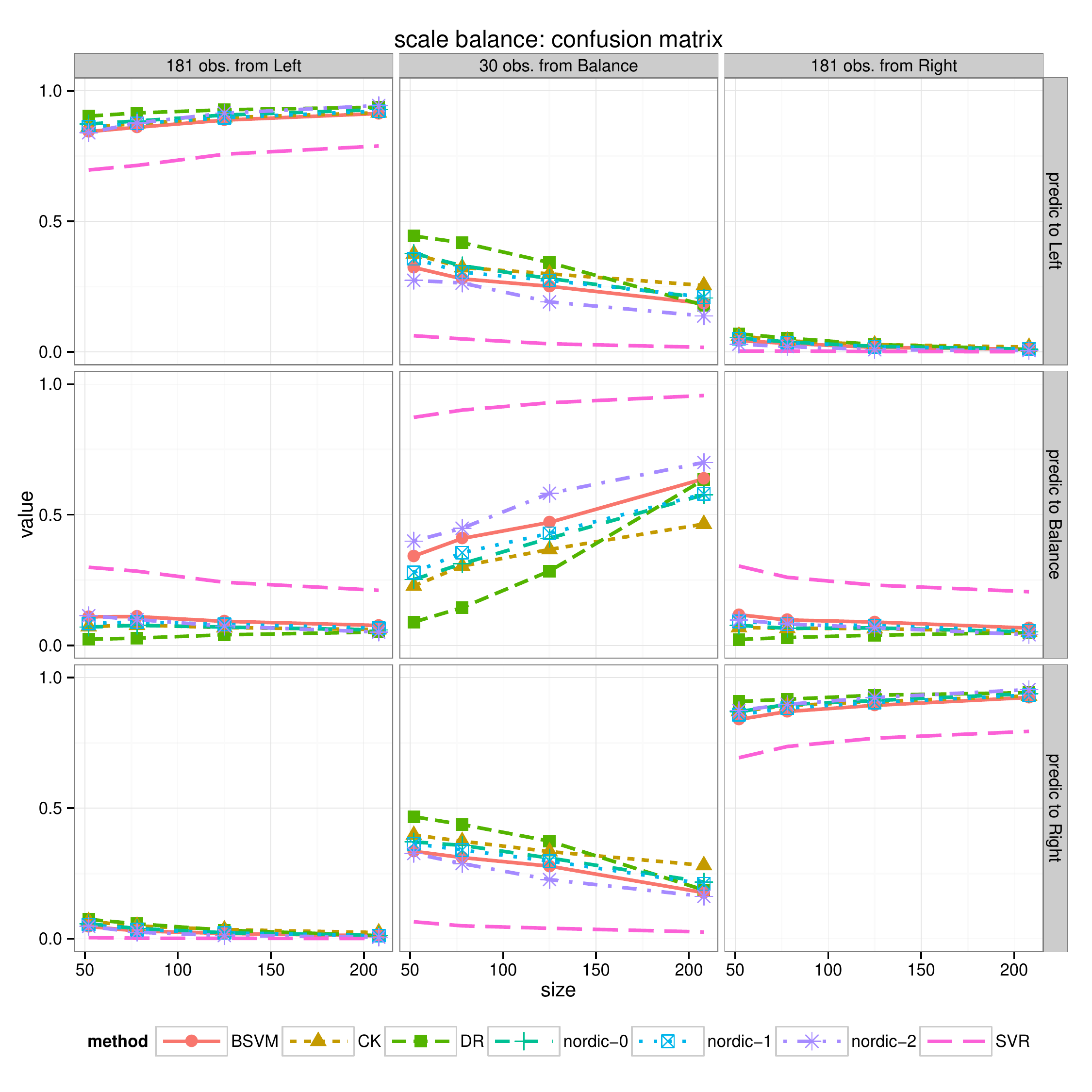}\vspace{-2em}
		\caption{\small Confusion matrices for the scale balance data set.}
	\label{fig:scale_conf}
\end{figure}

Figure \ref{fig:scale_conf} shows the confusion matrices. It can be seen that the poor performance of the SVR method is probably because it classifies much more instances to Class $B$, and this may be due to the arbitrary choice of the cut-off values $1.5$ and $2.5$. However, one may have no better way to choose the cut-offs except through another layer of tuning parameter selection. On the other hand, NORDIC-2 stands out as the best classifier due to its best performance on Class $B$ among the other methods (except for SVR.) Note that for Classes $L$ and $R$, all methods (except for SVR) perform more or less the same.

\section{Concluding Remarks}\label{sec:conclude}
In this article, three versions of NORDIC classifiers are proposed to make use of the order information in classifying ordinal data. All three classifiers train $(K-1)$ binary SVM classifiers simultaneously with extra constrains to ensure noncrossing among classification boundaries. The NORDIC-0 and NORDIC-1 methods focus on a sufficient condition for noncrossing and are solved by QP. The NORDIC-2 method aims for the exact condition for noncrossing but has to be solved by the  integer programming algorithm.

Let us turn our attention back to the formulation for NORDIC-0, (\ref{simpleNORDICformula})--(\ref{constraint_omega}). Without the additional constraints (\ref{constraint_b}) and (\ref{constraint_omega}), the NORDIC-0 method is the combination of $(K-1)$ independently trained SVM classifiers (with the common tuning parameter). It is known that for a single binary SVM classifier, the discriminant function is given by $f(\xv)=\sum_{j=1}^n\omega_{j}K(\xv_j,\xv)+b$. The coefficients $\omega_{i}=\alpha_{i}y_{i}$ is calculated by maximizing the following dual problem of SVM,
\begin{align}
\Lc_{SVM}&=\sum_i \alpha_i - \frac{1}{2}\sum_{i,j}\alpha_i\alpha_jy_iy_j K(\xv_i,\xv_j), \notag \\
\mb{subject to}~~~~ &\sum_i \alpha_i y_i =0,~\mb{and}~0\leq \alpha_i \leq C. \label{eq:sumtozero}
\end{align} See, for example, \citet{burges1998tutorial} for a tutorial.
The maximization problem above is the dual problem of SVM, while our NORDIC-0 method was based on the primal problem of SVM.  

One may wonder if a dual-based NORDIC is possible. Indeed, a variant of NORDIC can be viewed as to maximize the sum of $(K-1)$ such objective functions as in (\ref{eq:sumtozero}), with extra noncrossing constraints that $\omegav_{k\cdot}\geq \omegav_{(k+1)\cdot}$, that is $\alpha_{k,i}y_{i}^{(k)}\geq\alpha_{k+1,i}y_{i}^{(k+1)}$. Note that the constraints are the same as in NORDIC-0 but the objective function is based on the dual objective function. However, one can show that this formulation ultimately reduces to the method proposed by \citet{chu2005new}, namely, all the $(K-1)$ classifiers share the same $\omegav$ vector. Hence the CK method can be viewed as a special case in the NORDIC family. Note that in our NORDIC-0 proposal, we focus on the primal formulation. As a consequence, the resulting boundaries are not parallel to each other, leading to more flexibility.

The usefulness and efficiency of the proposed methods are supported by the comparison with the competitors. Promising results are obtained from simulated and data examples. Fisher consistency of the NORDIC method and asymptotic normality of the linear NORDIC method further validate the proposed methods.

There is a natural connection between ordinal classification and ordered logistic regression. Both methods fully utilize the ordinal class information. Their difference can be viewed as analogous to the difference between binary SVM and (binary) logistic regression, or that between multicategory SVM and multinomial logistic regression. It is interesting to explore the benefit of using machine learning techniques including NORDIC, over the modeling approaches such as ordered logistic regression. See \citet{Lee2015Does} for such a comparison in the binary case.

We have provided three distinct formulations. They may perform differently on different types of data sets, both in terms of the generalization error and the computational time; the derivation of these optimization problems may give insights into which kernels can more easily admit truly non-crossing boundaries. It is an interesting future research direction to identify specific kernels for which we can provide truly non-crossing boundaries.


\section*{Appendix}
\subsection*{Proof to Theorem \ref{thm1}}
Let $\hat\Zv_n$ and $\tilde\Zv_n$ denote $n^{1/2}\left\{(\hat\omegav_{k\cdot},\hat b_k)^T-(\omegav_{k\cdot}^0,b_k^0)^T\right\}$ and $n^{1/2}\left\{(\tilde\omegav_{k\cdot},\tilde b_k)^T-(\omegav_{k\cdot}^0,b_k^0)^T\right\}$, respectively. Then 
\begin{align*}
	&\left|\Pr\left(\hat\Zv_n\leq \uv\right)-\Pr\left(\tilde\Zv_n\leq \uv\right)\right|\\
	& = \left|\Pr\left(\hat\Zv_n\leq \uv\mid \hat\Zv_n\neq\tilde\Zv_n \right)-\Pr\left(\tilde\Zv_n\leq \uv\mid \hat\Zv_n\neq\tilde\Zv_n\right)\right|
	\\
	&\quad\times\Pr\left(\hat\Zv_n\neq\tilde\Zv_n\right)
\end{align*}
Since the first term in the product is bounded by 2, it suffices to show that $\Pr\left(\hat\Zv_n\neq\tilde\Zv_n\right)\rightarrow 0$. The event, $\hat\Zv_n\neq \tilde\Zv_n$, is equivalent to the event that the unconstrained binary linear SVM classifiers have boundaries crossing from each other, that is,
\begin{align*}
	n^{1/2}\left\{\sgn\left(\xv^T\tilde\omegav_{k\cdot}+\tilde b_{k}\right)-\sgn\left(\xv^T\tilde\omegav_{(k+1)\cdot}+\tilde b_{k+1}\right)\right\}< 0
\end{align*} for some $\xv\in\Sc$. This is only possible when $\xv^T\tilde\omegav_{k\cdot}+\tilde b_{k}<0$ and $\xv^T\tilde\omegav_{(k+1)\cdot}+\tilde b_{k+1}>0$. We consider their difference \begin{align*}
	n^{1/2}\left\{\left(\xv^T\tilde\omegav_{k\cdot}+\tilde b_{k}\right)-\left(\xv^T\tilde\omegav_{(k+1)\cdot}+\tilde b_{k+1}\right)\right\}.
\end{align*} This difference can be written as 
\begin{align*}
	&n^{1/2}	\left\{\left(\xv^T\tilde\omegav_{k\cdot}+\tilde b_{k}\right)-\left(\xv^T\omegav_{k\cdot}^0+ b_{k}^0\right)\right\}\\
	&\quad
	-n^{1/2}\left\{\left(\xv^T\tilde\omegav_{(k+1)\cdot}+\tilde b_{k}\right)-\left(\xv^T\omegav_{(k+1)\cdot}^0+ b_{k+1}^0\right)\right\}\\
	&\quad\quad+n^{1/2}\left\{\left(\xv^T\omegav_{k\cdot}^0+ b_{k}^0\right)-\left(\xv^T\omegav_{(k+1)\cdot}^0+ b_{k+1}^0\right)\right\}
\end{align*} 
Under the regularity conditions, and due to the results in \citet{koo2008bahadur}, the first two terms above are $O_p(1)$. Thus, $n^{1/2}\left\{\left(\xv^T\omegav_{k\cdot}^0+ b_{k}^0\right)-\left(\xv^T\omegav_{(k+1)\cdot}^0+ b_{k+1}^0\right)\right\}\leq -C<0$. This contradicts the fact that $\sgn\left(\xv^T\omegav_{k\cdot}^0+ b_{k}^0\right)\geq \sgn\left(\xv^T\omegav_{(k+1)\cdot}^0+ b_{k+1}^0\right)$ due to the assumption that the conditional density for each class is positive. Thus $\Pr\left(\hat\Zv_n\neq\tilde\Zv_n\right)\rightarrow 0$ which completes the proof.

\section*{Acknowledgements}
The work was partially supported by Binghamton University Harpur College Dean's New Faculty Start-up Funds and a collaboration grant from the Simons Foundation (\#246649 to Xingye Qiao). The author thanks the Statistical and Applied Mathematical Sciences Institute (SAMSI) for their generous support where he spent considerable amount of time when writing this article.

The author thanks Yichao Wu and Yufeng Liu for helpful comments. The author thanks Editor-in-Chief Prof. Heping Zhang and two anonymous referees for many thoughtful and constructive comments.
\bibliographystyle{asa}
\bibliography{nordic_references}
\end{document}